\documentclass{article}

\usepackage{booktabs}
\usepackage{placeins}
\usepackage{units}
\usepackage{colortbl}
\usepackage{mathtools}
\usepackage{multicol}
\usepackage{graphicx}
\usepackage{numprint}

\usepackage{color,graphics,epsfig,latexsym,amssymb,amsmath,bm}
\usepackage{graphicx}

\usepackage{listings}
\usepackage{tabularx}
\usepackage[english]{babel}
\usepackage{gensymb}
\usepackage{epstopdf}
\usepackage{pgfplots, pgfplotstable}
\usepackage{tikz-timing}
\usepackage{subcaption}

\usepackage{xspace}

\usepackage{url}
\usepackage{color}
\usepackage[binary-units=true]{siunitx}
\usepackage{listings}
\usepackage{dirtree}
\usepackage{IEEEtrantools}

\usepackage{cprotect}

\usepackage{mwe,tikz}\usepackage[percent]{overpic}
\usepackage{multirow}
\usepackage{subcaption}
\usepackage{enumitem}

\usepackage[rightcaption]{sidecap}

\usepackage{textcomp}
\usepackage{gensymb}
\usepackage{nameref,zref-user, zref-xr}
\usepackage{hyperref}

\hypersetup{
    colorlinks,
    linkcolor={blue!30!black},
    citecolor={blue!30!black},
    urlcolor={blue!30!black}
}

\usepackage{acronym}

\newif\ifMaxQuality

\usepackage[acronym]{glossaries}
\newacronym{DOF}{DoF}{Degree of Freedom}
\newacronym{GPS}{GPS}{Global Positioning System}
\newacronym{UTM}{UTM}{Universal Transverse Mercator}
\newacronym{TI}{TI}{Thermal-Infrared}
\newacronym{LWIR}{LWIR}{Long-Wave Infrared}
\newacronym{NIR}{NIR}{Near-Infrared}
\newacronym{UAV}{UAV}{Unmanned Aerial Vehicle}
\newacronym{CNN}{CNN}{Convolutional Neural Network}
\newacronym{HOG}{HOG}{Histogram of Oriented Gradient}
\newacronym{SVM}{SVM}{Support Vector Machine}
\newacronym{FPN}{FPN}{Feature Pyramid Networks}
\newacronym{SSD}{SSD}{Single Shot Detector}
\newacronym{PDF}{PDF}{Probability Density Function}
\newacronym{SR}{SR}{Systematic Resampling}
\newacronym{PF}{PF}{Particle Filter}
\newacronym{IMU}{IMU}{Inertial Measurement Unit}
\newacronym{IOU}{IoU}{Intersection over Union}
\newacronym{SAR}{SaR}{Search and Rescue}
\newacronym{SWE}{SWE}{Sliding Window Estimator}
\newacronym{JPL}{JPL}{Jet Propulsion Laboratory}
\newacronym{ASL}{ASL}{Autonomous Systems Lab}
\newacronym{ETHZ}{ETHZ}{ETH Zurich}
\newacronym{V4R}{V4R}{Vision for Robotics Lab}
\newacronym{CVG}{CVG}{Computer Vision and Geometry Group}
\newacronym{LiDAR}{LiDAR}{Light Detection And Ranging}
\newacronym{SLAM}{SLAM}{Simultaneous Localization And Mapping}
\newacronym{GNSS}{GNSS}{Global Navigation Satellite System}
\newacronym{DL}{DL}{Deep Learning}
\newacronym{ML}{ML}{Machine Learning}
\newacronym{ICLK}{ICLK}{Inverse Compositional Lucas-Kanade}
\newacronym{DNN}{DNN}{Deep Neural Network}

\newacronym{MDN}{MDN}{Mixture Density Networks}
\newacronym{PVV}{PVV}{Probabilistic Vertex Voting}
\newacronym{AS}{AS}{Active Search}
\newacronym{VAV}{VAV}{Vote-and-verify}
\newacronym{VLAD}{VLAD}{Vector of Locally Aggregated Descriptors}
\newacronym{BOW}{BOW}{Bag-of-Words}
\newacronym{MLE}{MLE}{Maximum Likelihood Estimation}
\newacronym{mm-ISAM}{mm-ISAM}{multi-modal Incremental Smoothing and Mapping}
\newacronym{DEM}{DEM}{Digital Elevation Map}
\newacronym{VI}{VI}{Visual-Inertial}
\newacronym{MI}{MI}{Mutual Information}
\newacronym{SO}{SO}{Ordnance Survey}
\newacronym{ICP}{ICP}{Incremental Closest Point}
\newacronym{kNN}{kNN}{k-Nearest Neighbors}
\newacronym{PnP}{PnP}{Perspective-n-Point}
\newacronym{RANSAC}{RANSAC}{Random Sample Consensus}
\newacronym{GPU}{GPU}{Graphics Processing Unit}

\usepackage{booktabs}

\title{SD-6DoF-ICLK: Sparse and Deep Inverse Compositional Lucas-Kanade Algorithm on SE(3)}
\author{Timo Hinzmann\thanks{All authors are with the Autonomous Systems Lab, ETH Zurich.} \and Roland Siegwart\footnotemark[1]}

\date{}

\begin{document}

%
%
%
\maketitle

\begin{abstract}
This paper introduces SD-6DoF-ICLK, a learning-based Inverse Compositional Lucas-Kanade (ICLK) pipeline that uses sparse depth information to optimize the relative pose that best aligns two images on SE(3). %
To compute this six Degrees-of-Freedom (DoF) relative transformation, the proposed formulation requires only sparse depth information in one of the images, which is often the only available depth source in visual-inertial odometry or Simultaneous Localization and Mapping (SLAM) pipelines.
In an optional subsequent step, the framework further refines feature locations and the relative pose using individual feature alignment and bundle adjustment for pose and structure re-alignment.
The resulting sparse point correspondences with subpixel-accuracy and refined relative pose can be used for depth map generation, or the image alignment module can be embedded in an odometry or mapping framework.
Experiments with rendered imagery show that the forward SD-6DoF-ICLK runs at $\unit[145]{ms}$ per image pair with a resolution of $\unit[752\times480]{pixels}$ each, and vastly outperforms the classical, sparse 6DoF-ICLK algorithm, making it the ideal framework for robust image alignment under severe conditions.
\end{abstract}

\section{Introduction}
Robust image alignment under challenging conditions is an important core capability towards safe autonomous navigation of robots in unknown environments.
This paper focuses primarily on the \gls*{ICLK}~\cite{lucas1981iterative, baker2004lucas} algorithm that optimizes the alignment between two images on SE(3) by utilizing dense or sparse depth information.
One advantage of this approach in contrast to feature-based alignment is that a costly outlier rejection step can be avoided.
Applications of six-\gls*{DOF}-\gls*{ICLK} include rigid stereo extrinsics refinement after shocks, visual-inertial odometry systems \cite{Forster2017}, or non-rigid stereo pair tracking \cite{Hinzmann2019}.
In this paper, the 6DoF-\gls*{ICLK} is leveraged with the help of \gls*{DL} to make parameters in the framework trainable and the image alignment robust against e.g., challenging light conditions.
%

%

\section{Related Work}
Image alignment approaches can be divided into feature-based and direct methods.
Feature-based methods have come a long way from computationally expensive hand-crafted feature detectors and descriptors (SIFT \cite{Lowe:IJCV2004}, SURF \cite{Bay2006}) to faster binary variants (BRISK \cite{leutenegger2011brisk}) and recently trainable approaches (SuperPoint \cite{detone2017superpoint}, D2-Net \cite{Dusmanu_2019_CVPR}). 
Likewise, direct methods have seen many variants and may rely on dense (DTAM \cite{Newcombe2011}) or sparse depth information (SVO \cite{Forster2017}), Mutual Information based Lucas-Kanade tracking \cite{DowsonB08}, and more recently \gls*{DL}-variants of \gls*{ICLK} \cite{Lv19cvpr}.
%
In this context, we propose SD-6DoF-\gls*{ICLK}, a learning-based sparse Inverse Compositional Lucas-Kanade (ICLK) algorithm for image alignment on SE(3) using sparse depth estimates.
The implemented SD-6DoF-\gls*{ICLK} algorithm is optimized for \gls*{GPU} operations to speed up batch-wise training.
The output of the framework consists of sparse feature pairs in both images and the relative pose connecting the two cameras.
The feature locations and the estimated relative pose can be further refined using an individual feature alignment step with a subsequent pose, and optional structure refinement, as proposed in \cite{Forster2017}.
For depth map generation, the estimated relative pose can be fed to classical rectification and depth estimation modules, or to \gls*{DL}-based depth from multi-view stereo algorithms.
%

%
\section{Methodology}
\begin{figure}[htb]
\includegraphics[width=\linewidth]{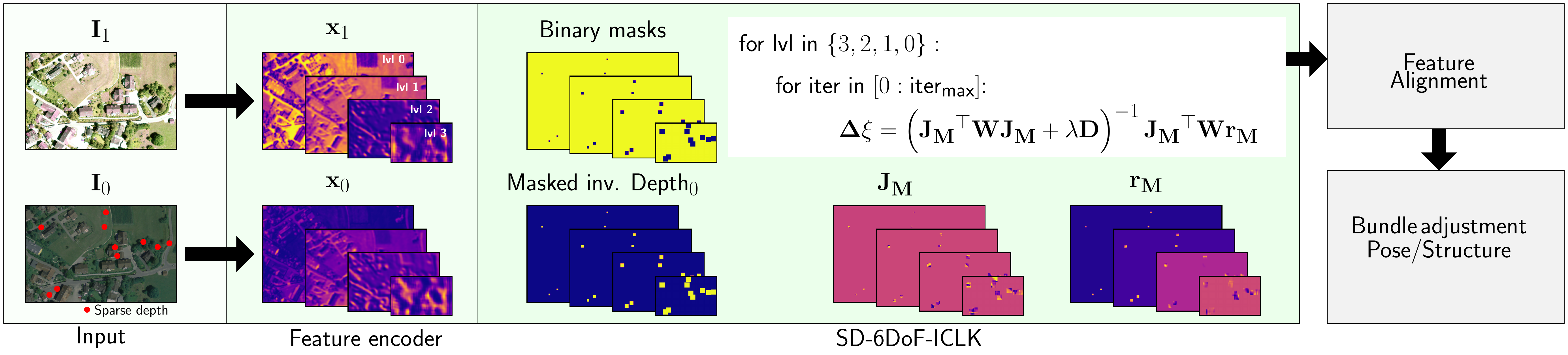}
\caption{Overview of SD-6DoF-ICLK with feature, pose, and structure refinement.}
\label{fig:sparse:overview}
\end{figure}

Our proposed SD-6DoF-\gls*{ICLK} framework is depicted in Fig.~\ref{fig:sparse:overview}:
The input is a grayscale or colored reference image $\mathbf{I}_0$ with sparse features.
For every sparse feature, a depth estimate is assumed to be known.
The objective is to find the relative 6DoF pose $\smash{\mathbf{T}^{C_1}_{C_0}}$ that best aligns the reference image $\mathbf{I}_0$ to $\mathbf{I}_1$, where $\mathbf{I}_1$ may also be grayscale or colored but contains no depth information or extracted features.
To align the images on SE(3) with the described input data, we adopt \cite{Lv19cvpr} to sparse depth information as follows:
The input images $\mathbf{I}_0$ and $\mathbf{I}_1$ are color normalized and fed as single views to the feature encoder described in \cite{Lv19cvpr}.
This operation returns four pyramidal images with resolution $752\times480$ (level 0, input image resolution), $376\times240$, $188\times120$ and $94\times60$ (level 3).
The sparse image alignment algorithm described in \cite{Forster2017} is designed for CPU operations and explicitly iterates over the pixels of every patch.
Instead, we formulate the problem with binary masks to exploit the full advantage of indexing with \glspl*{GPU} and to allow fast batch-wise training and inference.
To achieve this, binary masks are created at every level with a patch size of $8\times8$ pixels surrounding the sparse feature.
Similarly, the sparse inverse depth image for image $\mathbf{I}_0$ is generated by setting every pixel within the patch to the inverse depth value.

Fig.~\ref{fig:sparse:overview} shows the pseudo-code of the inverse compositional algorithm that, starting from the highest level, iterates over all pyramidal images.
For every pyramidal image, the warp parameters are optimized using the Levenberg-Marquardt update step \cite{marquardt:1963, Lv19cvpr}:
\begin{align}
\mathbf{\Delta \xi} = \left(\mathbf{J_M}^\top\mathbf{W}\mathbf{J_M}+\lambda \mathbf{D}\right)^{-1}\mathbf{J_M}^\top\mathbf{W}\mathbf{r_M}
\end{align}
where $\mathbf{J_M}$ and $\mathbf{r_M}$ denote the Jacobian $\mathbf{J}$ and residual $\mathbf{r}$ after applying the binary mask of the corresponding pyramidal layer.
The convolutional M-Estimator proposed in \cite{Lv19cvpr} is used to learn the weights in $\mathbf{W}$.
The damping term $\lambda$ is set to $\smash{1e^{-6}}$ throughout the experiments.

As shown in Fig.~\ref{fig:sparse:overview}, the framework continues with a feature alignment step and bundle adjustment step to achieve subpixel accuracy \cite{Forster2017}.

\section{Experiments and Results}
\paragraph{Implementation}
The SD-6DoF-ICLK algorithm and feature alignment step is implemented in pytorch~\cite{NEURIPS2019_9015} and adopted from~\cite{Lv19cvpr,Forster2017}.
The pose is then refined with GTSAM~\cite{dellaert2012factor} using a Cauchy loss, to reduce the influence of outliers, and Levenberg-Marquardt for optimization.
\paragraph{Datasets}
To generate a large amount of training, validation, and test data we implemented a shader program in OpenGL~\cite{woo1999opengl} that takes an orthomosaic and elevation map as input and renders an RGB and depth image given a geo-referenced pose $\mathbf{T}^W_C$, camera intrinsics matrix $\mathbf{K}$, and distortion parameters \cite{Hinzmann2020_loc}.
No distortion parameters are set in this paper as the \gls*{ICLK} algorithm expects undistorted images as input.
Camera positions and orientations of pose pairs are uniformly sampled from locations above the orthomosaic and rejected if the camera's field of view is facing areas outside the map.
Training and test data are rendered from two different satellite orthomosaics selected from nearby but non-overlapping locations. 
The validation set\footnote{The validation split is set to 20\% of the total set dedicated for training and validation.} is drawn from the shuffled training data.
The OpenGL renderer's task is to augment the training and test data geometrically.
Appearance variations are generated using pytorch's build-in color-jitter functionality that randomly sets brightness, contrast, saturation, and hue.
This is to emulate challenging light conditions like over-exposure or lens flares.
This augmentation creates $10$ new, color-altered images for every inserted original image.
Sparse features are drawn randomly from a uniform distribution such that $\mathbf{p}=[\mathcal{U}(b,W - b); \mathcal{U}(b,H - b)])$ with border $b=2\cdot P + 1$, patch width $P=8$, image width $W$ and image height $H$.
The number of sparse features for all training, validation, and test set is set to $50$.

\begin{figure}[htb]
\centering
\begin{minipage}[b]{0.5\textwidth}
\vspace{0pt}
\centering
\includegraphics[width=\linewidth]{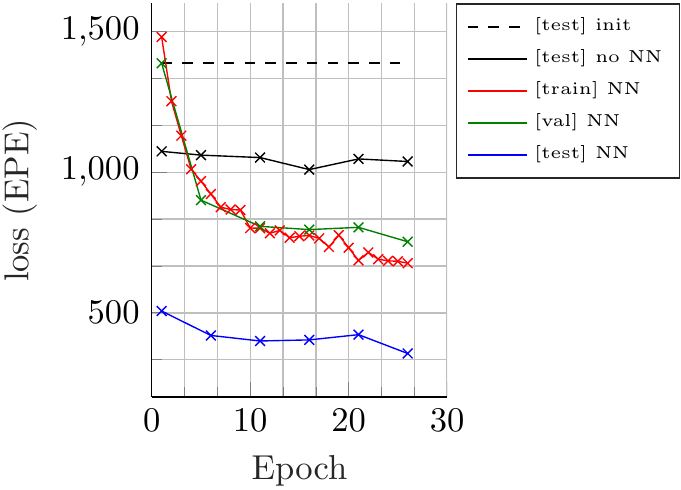}
\caption{Training, validation and test loss (3D End-Point-Error EPE) over $26$ epochs.}
\label{fig:train}
\end{minipage}
\hfill
\begin{minipage}[b]{0.45\textwidth}
\vspace{0pt}
\captionsetup{type=table} 
\resizebox{\textwidth}{!}{\begin{tabular}{l|l}
\textbf{Param} & \textbf{Value} \\ \hline
Batch size & $4$ (max. memory)\\
Epochs & $26$\\
Num. images train/val & $\numprint{20000}$\\
Validation split & $0.2$ \\
Num. images test & $800$\\
Input image resolution& $752\times 480$ \\
Max. iterations & 10 \\
Optimizer & SGD \\
Initial learning rate& $\smash{1.0e^{-5}}$ \\
Momentum& $0.9$ \\
Nesterov& False \\
Weight decay& $\smash{4e^{-4}}$ \\
Learning rate decay epochs& $[5,10,20]$ \\
Learning rate decay ratio& $0.5$ \\
Validation frequency& $5$ \\
Test frequency& $5$ \\
\end{tabular}}
\caption{Training, validation, and test parameters used for session visualized in Fig.~\ref{fig:train}.}
\label{tab:train}
\end{minipage}
\end{figure}
\begin{figure}[htb]
\begin{multicols}{4}
    \includegraphics[width=\linewidth]{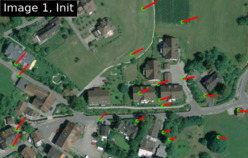}\par
    \includegraphics[width=\linewidth]{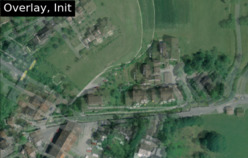}\par
    
    \includegraphics[width=\linewidth]{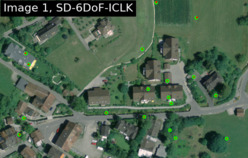}\par
    \includegraphics[width=\linewidth]{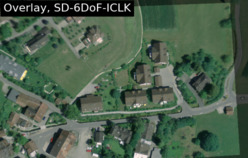}\par

    \includegraphics[width=\linewidth]{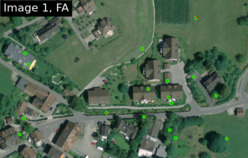}\par
    \includegraphics[width=\linewidth]{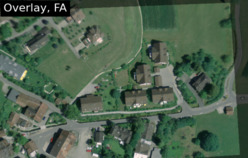}\par

    \includegraphics[width=\linewidth]{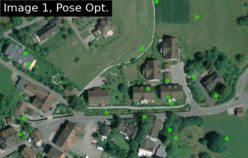}\par
    \includegraphics[width=\linewidth]{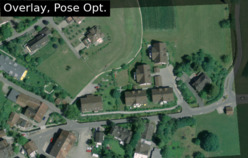}
\end{multicols}
\caption{Qualitative results of the image alignment framework.}
\label{fig:sparse:qualitative}
\end{figure}

\paragraph{Results}
Fig.~\ref{fig:train} presents the results from a training, validation, and test session over $26$ epochs.
Training parameters are listed in Tab.~\ref{tab:train}.
Analogue to \cite{Lv19cvpr}, the utilized training loss is the 3D End-Point-Error (EPE) using the rendered dense depth map.
After epoch $10$, the training loss (red) is below the validation loss (green) and continues to decrease.
As the appearance of the images is randomly changed, it may occur that the validation is below the training loss as seen here in the initial set of epochs.
On every fifth epoch, the test set is evaluated.
The loss of the test set given the initial, incorrect relative transformation is illustrated by the black dashed line for reference.
The solid black line is the test set evaluated for 6DoF-ICLK without learning (also max.\ $10$ iterations) and returns roughly the same result independent of the epoch as expected.
Evaluating the test set with SD-6DoF-ICLK demonstrates that the classical 6DoF-ICLK is clearly outperformed.
%
%
%
Tab.~\ref{tab:sparse:quantitative} shows the pixel (euclidean distance), translational, and rotational errors that continuously decrease for the subsequent image alignment steps.
These results are visualized in Fig.~\ref{fig:sparse:qualitative}:
The first row shows $\mathbf{I}_1$ with the features projected from $\mathbf{I}_0$ based on the current estimate for the relative transformation $\smash{\mathbf{T}^{C_1}_{C_0}}$.
The red lines illustrate the error with respect to the ground truth feature location.
Given the estimate for $\smash{\mathbf{T}^{C_1}_{C_0}}$, $\mathbf{I}_1$ can be overlaid over $\mathbf{I}_0$, which is shown in the second row.
\begin{table}
\resizebox{\linewidth}{!}{\begin{tabular}{l|c|c|c|c}
 &  \textbf{Initial} & \textbf{SD-6DoF-ICLK} & \textbf{Feature Alignment} & \textbf{Pose Optimization} \\ \hline
$e_\text{pixel}$ & 33.986 & 1.286 & 0.413 & 0.121 \\
$e_\text{transl.}$ & 4.934 & 3.257 & 3.257 & 0.089 \\
$e_\text{rot.}$ &  0.075 & 0.020 & 0.020 & 0.000 \\
\end{tabular}}
\caption{Pixel, translational, and rotational error for the subsequent image alignment steps.}
\label{tab:sparse:quantitative}
\end{table}

\setlength{\columnsep}{2pt}
\paragraph{Runtime}
The inference time of the SD-6DoF-ICLK is $\unit[145]{ms}$ on a GeForce RTX 2080 Ti (12GB) for an image resolution of $752\times 480$ pixels, four pyramidal layers, and a maximum of $10$ iterations of the incremental optimization.
Note that the image resolution of $752\times 480$ was selected for training and forward inference, as it represents the target camera resolution on our \gls*{UAV}.
A smaller resolution, however, could be set to speed up the training process and to increase the batch size if desired.


\section{Conclusion}
This work introduced SD-6DoF-ICLK, a learning-based sparse Inverse Compositional Lucas-Kanade (ICLK) algorithm, enabling robust image alignment on SE(3) with sparse depth information as input, and optimized for \gls*{GPU} operations.
A synthetic dataset rendered with OpenGL shows that SD-6DoF-ICLK outperforms the classical sparse 6DoF-ICLK algorithm by a large margin, making the proposed algorithm an ideal choice for robust image alignment.
The proposed SD-6Dof-ICLK is able to perform inference in $\unit[145]{ms}$ on a GeForce RTX 2080 Ti with input images at a resolution of $752\times 480$ pixels.
To further refine the feature locations and relative pose, individual feature alignment with subsequent pose and, if required, structure refinement is applied as proposed in \cite{Forster2017}.
In future work, the framework could be embedded into an odometry or mapping framework, or used for depth image generation.
\section*{Acknowledgments}
The authors thank \emph{Geodaten \textsuperscript{\textcopyright{}} swisstopo} for access to the satellite imagery.
%

\FloatBarrier
\addcontentsline{toc}{chapter}{Bibliography}

\bibliography{lib.bib}
\bibliographystyle{abbrv}

\end{document}